\documentclass[12pt]{article}
\usepackage{amssymb}
\usepackage{amsfonts}
\usepackage{amsmath}
\usepackage[pdftex]{graphicx}
\usepackage{setspace} 
\usepackage[all]{xy}
\begin{document}

\title{moco: Fast Motion Correction for Calcium Imaging}

\author{Alexander Dubbs, James Guevara, \\ Darcy S. Peterka, Rafael Yuste}

\maketitle

\begin{abstract}

Motion correction is the first in a pipeline of algorithms to analyze calcium imaging videos and extract biologically relevant information, for example the network structure of the neurons therein. Fast motion correction would be especially critical for closed-loop activity triggered stimulation experiments, where accurate detection and targeting of specific cells in necessary. Our algorithm uses a Fourier-transform approach, and its efficiency derives from a combination of judicious downsampling and the accelerated computation of many $L_2$ norms using dynamic programming and two-dimensional, fft-accelerated convolutions. Its accuracy is comparable to that of established community-used algorithms, and it is more stable to large translational motions. It is programmed in Java and is compatible with ImageJ.
\end{abstract}

\section{Introduction}

Calcium imaging, first used to measure the activity of neurons in the early 1990s \cite{Yuste1991}, has been successfully applied throughout the nervous system.  It allows us to see the behavior of neurons in awake behaving mice, using either chemical or genetic calcium indicators, with confocal microscopy, two-photon microscopy, or wearable imaging devices \cite{Grienberger2012}. As a result, it is an increasingly useful tool for identifying the neural substrate of mouse behaviors. However, calcium imaging videos have difficult noise properties, including white noise and motion artifacts which must be corrected in a preprocessing step before proper analysis can be undertaken.

Motion correction is the first step in the analysis of calcium images. After they are motion-corrected, ROIs are identified, and time-activity graphs are made from each ROI. If the motion-correction is low-quality, then the time-activity graphs suffer, and the reconstructed networks may have errors. For real time closed loop operation, if the motion correction is slow, it cannot be done while the mouse is in the microscope, and the experiment fails.

TurboReg \cite{Thevanaz1998} is a commonly used algorithm to do motion correction. It uses a downsampling strategy, which is prerequisite for speed, and it uses a template image, which is necessary for accuracy. We have developed a similar method, called {\it moco} (MOtion COrrector), which adopted both strategies, since correcting one image against the next in the stack results in unacceptable roundoff errors. Other approaches use HMMs \cite{Collman2010}, \cite{Kaifosh2014} or other techniques \cite{Guizar2008}, \cite{Li2008}, \cite{Greenberg2009}, \cite{Poole2015}, \cite{Ringach}. \cite{Guizar2008} is the only one similar mathematically to, and may be slightly faster than, moco, but it has accuracy problems (see Figure 2).

moco uses downsampling and a template image, and it can be called from ImageJ. However, it is faster than TurboReg \cite{Thevanaz1998} at translation-based motion correction because it uses dynamic programming and two-dimensional fft-acceleration of two-dimensional convolutions. \cite{Guizar2008} also uses the fft approach but uses a different objective function that does not require dynamic programming; we believe that our approach is more robust to corrupted data, see Figure 2. Image Stabilizer is as fast for small images, but is very slow for standard-size images. Running on our own datasets, moco appears faster than all approaches compatible with ImageJ.

moco corrects every image in the video by comparing every possible translation of it with the template image, and chooses the one which minimizes the $L_2$ norm of the difference between the images in the overlapping region, $D$, divided by the area of $D$. The fact that it is so thorough makes it robust to long translations in the data. More complicated non-translation image warps are usually unnecessary for fixing calcium images, which suffer from spurious translations, which moco corrects, and spurious motion in the Z-direction, which is very hard to correct. Our approach also uses cache-aware upsampling: when an image is aligned with the template in the downsampled space, it must be jittered when it is upsampled to see which jitter best aligns with the upsampled template. We do this in such a way that data that is used recently is reused immediately, making the implementation faster. Hence, moco is an efficient motion correction of calcium images, and is likely to become a useful tool for processing calcium imaging movies.

\section{Mathematical Development}

Let $a_{i,j}$, for $i = 1,\ldots,m$ and $j = 1,\ldots,n$ be an image in the stack. We assume $a$ is downsampled if it is larger than $256\times 256$. Let $b_{i,j}$ be the template image against which to align $a$. We want to pick $(s,t)$ such that $\max(|s|,|t|) < w$, where $w$ is input by the user, and
$$ f_{s,t} = \frac{1}{{\rm Area}(D_{s,t})}\sum_{(i,j)\in D_{s,t}}(a_{i+s,j+t}-b_{i,j})^2 $$
is minimal, where $D_{s,t}$ is the set of ordered pairs of integers $(i',j')$ such that $1\leq i' \leq m$, $1\leq j' \leq n$, $1\leq i'+s\leq m$, and $1\leq j'+t\leq n$. If we do this for every image $a$ in the stack, we have then motion corrected the video, and we are done, up to a short upsampling step. To upsample, multiply the optimal $(s,t)$ by $2$ and do a local search to minimize $f_{s,t}$ on the finer grid. Now,
$$ {\rm Area}(D_{s,t})f_{s,t} = \sum_{(i,j)\in D_{s,t}}\hskip -.1in a_{i+s,j+t}^2 + \sum_{(i,j)\in D_{s,t}}\hskip -.1in b_{i,j}^2 - 2\hskip -.1in\sum_{(i,j)\in D_{s,t}}\hskip -.1in a_{i+s,j+t}b_{i,j}.$$
The first two sums can be computed via dynamic programming. Let's consider $a$ when $s$ and $t$ are negative. Let
$$ g_{s,t} =  \sum_{(i,j)\in D_{s,t}}\hskip -.1in a_{i+s,j+t}^2. $$
We have that
$$ g_{s,t} = g_{s-1,t} + g_{s,t-1} - g_{s-1,t-1} + a_{m+s,n+t}^2. $$
Hence, the first two sums can be computed for all $(s,t)$ in $O(mn)$ time, which is unaffected by a constant amount of downsampling. It suffices to compute for all $(s,t)$ such that $\max(|s|,|t|) < w$,
$$ h_{s,t} = \sum_{(i,j)\in D_{s,t}}\hskip -.1in a_{i+s,j+t}b_{i,j}. $$
Let $\hat{b}$ be $b$ rotated $180$ degrees. Using MATLAB notation, let
$$ \tilde{a} =  {\rm fft2}([[a,{\rm zeros(m,w)}];{\rm zeros(w,n+w)}]), $$
$$ \tilde{b} = {\rm fft2}([[\hat{b},{\rm zeros(m,w)}];{\rm zeros(w,n+w)}]). $$
Commas denote horizontal concatenation, semicolons denote vertical concatenation, and ${\rm zeros}(x,y)$ is an $x\times y$ matrix of zeros. For equally sized matrices $X$, $Y$, let $Z = X\odot Y$ mean $Z_{i,j} = X_{i,j}Y_{i,j}$. Then
$$ {\rm ifft2}(\tilde{a}\odot\tilde{b}) $$
is a rearrangement of $h$. Since ${\rm fft2}$'s are fast, that means $h$ can be computed for all $(s,t)$ in $O(mn\log(mn))$ time. Hence, after upsampling, the entire video can be aligned in $O(mnT\log(mn))$ time, where $T$ is the number of slides in the video.

After $(s,t)$ are chosen to minimize $f_{s,t}$, they are multiplied by two multiple times to upsample. Every time they are multiplied by $2$, $f_{2s+u,2t+v}$ are computed for $u,v\in\{0,-1,1\}$ to see which $u$ and $v$ are minimal. These nine evaluations of $f$ are done with a cache-aware algorithm for speed.

\section{Results}

We compare moco in speed to TurboReg \cite{Thevanaz1998} on its translation mode, using both the ``fast'' and ``accurate'' settings. We also compare it to Image Stabilizer using its default settings \cite{Li2008} (it can be made faster by changing the settings but the accuracy is poor). We use several real calcium imaging videos, which we say are $m\times n\times T$ if they contain $T$ slides of size $m\times n$. If the images are larger than $256\times 256$, we downsample once, otherwise, we do not downsample. We have found that dowsampling $3$ and $4$ times causes severe errors so we avoid those settings. In addition, we have compared moco to TurboReg on synthetic images with severe translational motion artifacts and have found that moco is slightly more accurate. All times are in seconds. The template used for every video is the first image in the video for both moco and TurboReg. moco uses a maximum translation width of $\min(m,n)/3$ in both the $i$ and $j$ directions.
\vskip .2in
\begin{center}
\begin{tabular}{|c|c|c|c|c|}
\hline
Size & moco & TurboReg & TurboReg (slow) & Image Stabilizer\\
\hline
$512\times 512\times 1500$ & 66{\rm s} & 110{\rm s} & 242{\rm s} & 304{\rm s} \\
\hline
$512\times 512\times 2000$ & 90{\rm s} & 170{\rm s} & 298{\rm s} & 464{\rm s} \\
\hline
$512\times 512\times 6984$ & 288{\rm s} & 632{\rm s} & 1303{\rm s} & 2277{\rm s} \\
\hline
$416\times 460\times 1000$ & 35{\rm s} & 71{\rm s} & 132{\rm s} & 41{\rm s} \\
\hline
$256\times 256\times 2028$ & 84{\rm s} & 121{\rm s} & 154{\rm s} & 34{\rm s} \\
\hline
\end{tabular}
\end{center}
\vskip .2in

\noindent As is clear from the table, moco is faster than its most used current method, TurboReg. It may be marginally slower than \cite{Guizar2008}, but Figure 2 proves that a code we have created to have similar results to \cite{Guizar2008} is inaccurate. Figure 1 shows the first two images of a corrupted video on the first row. moco corrections are on the second row. It is clear that moco can fix the image, even though the problems with it are severe. Figure 2 shows the mean image from a corrupted video, and the mean image of moco and TurboReg corrections, as well as the correction from our MATLAB version of the \cite{Guizar2008} algorithm.

\begin{figure}
\begin{centering}
\rule{.4in}{0in}\begin{tabular}{cc}
\includegraphics[scale=1.5]{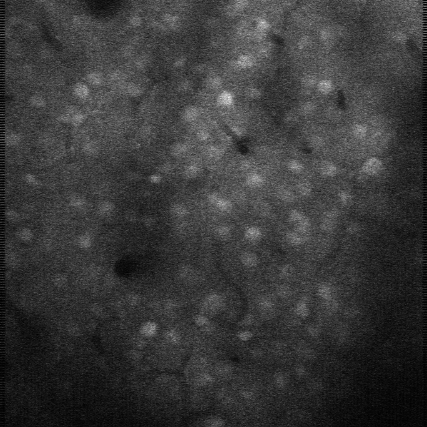} & \includegraphics[scale=1.5]{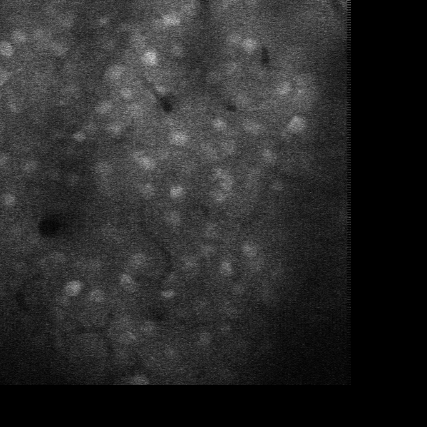} \\
 1.a. & 2.a. \\
\includegraphics[scale=1.5]{FirstPre.pdf} & \includegraphics[scale=1.5]{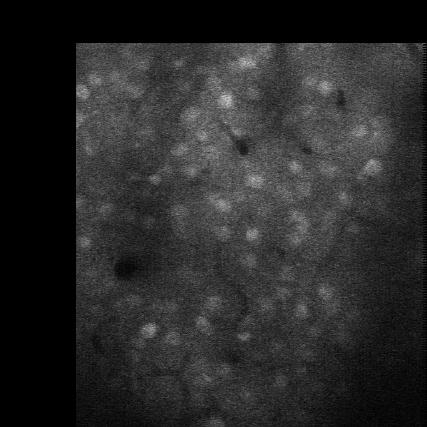} \\
1.b. & 2.b. \\
\end{tabular}
\end{centering}
\caption{Images are $317.44\mu m\times 317.44\mu m$. 1.a. and 2.a. are the first two images of a long, badly corrupted video submitted to moco. 1.b. and 2.b. are the first two corrected images. One can see that 1.a. and 1.b. are the same, since 1.a. is used as the template image. However, 2.b. is moved, so that if it is moved to the left to overlap 1.b., it matches it almost perfectly except where it is black.}
\end{figure}

\begin{figure}
\begin{centering}
\rule{.4in}{0in}\begin{tabular}{cc}
\includegraphics[scale=1.5]{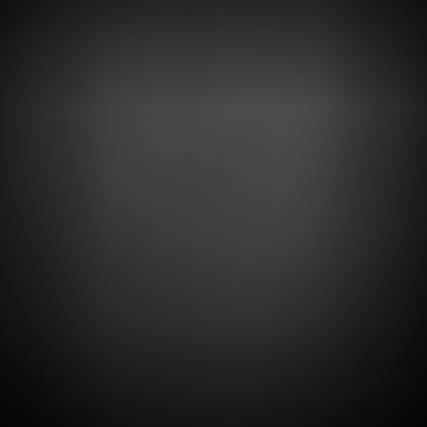} & \includegraphics[scale=1.5]{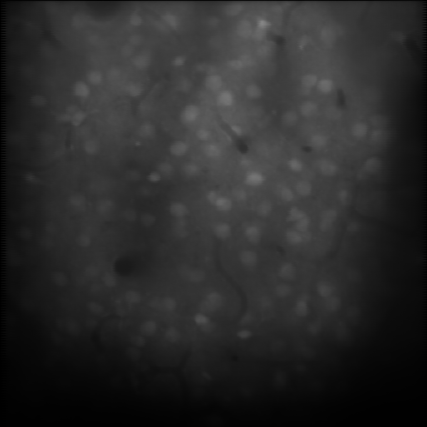} \\
 1.a. & 2.a. \\
\includegraphics[scale=1.5]{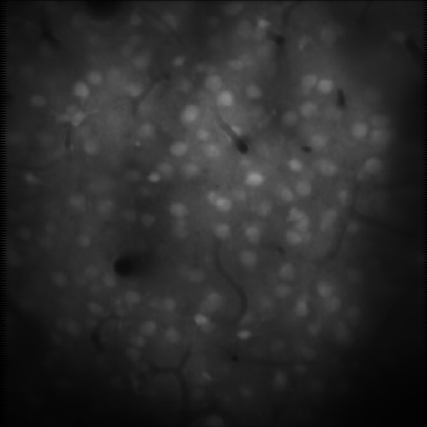} & \includegraphics[scale=1.5]{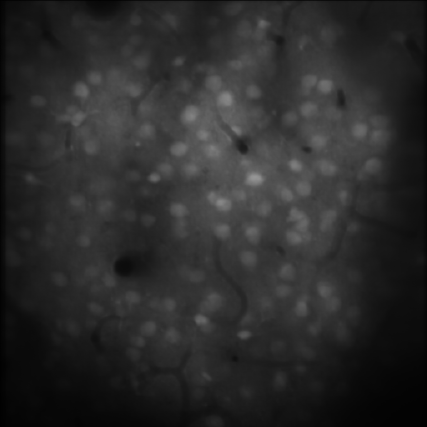} \\
1.b. & 2.b. \\
\includegraphics[scale=1.5]{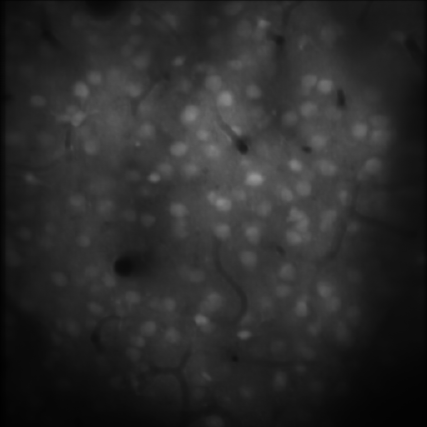} & \includegraphics[scale=1.5]{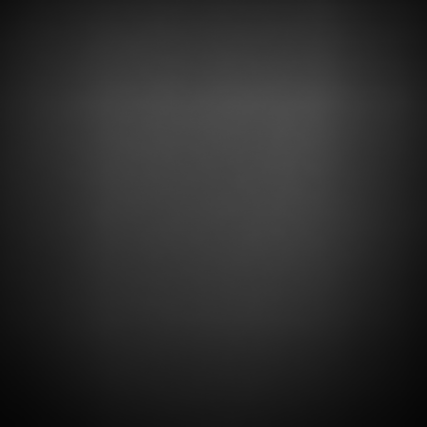} \\
1.c. & 2.c. \\
\end{tabular}
\end{centering}
\caption{Images are $317.44\mu m\times 317.44\mu m$. 1.a. is the mean of every image in a badly corrupted video. 2.a. is the mean of the corrected video using our implementation of the \cite{Guizar2008} approach. 1.b. is the mean of the corrected video using moco. 2.b. is the mean of the corrected video using TurboReg (accurate mode), \cite{Thevanaz1998} 1.c. is the mean of the corrected video using TurboReg (accurate mode). 2.c. is the mean of the corrected video using Image Stabilizer. We see that moco and TurboReg have superior performance.}
\end{figure}

\section{Acknowledgments}

The first author would like to thank Julia Sable for her help, and Inbal Ayzenshtat, Jesse Jackson, Jae-eun Miller, Luis Reid, Weijian Yang, and Weiqun Fang for their datasets. The first author would also like to thank the Columbia University Data Science Institute for their ongoing support. This research is supported by the NEI (DP1EY024503, R01EY011787), NIHM (R01MH101218, R41MH100895) and DARPA contract W91NF-14-1-0269. This material is based upon work supported by, or in part by, the U. S. Army Research Laboratory and the U. S. Army Research Office under contract number W911NF-12-1-0594 (MURI).

\end{document}